\documentclass[preprint,12pt,authoryear]{elsarticle}

\usepackage{amssymb}
\usepackage{amsmath}

\usepackage{subcaption}
\usepackage{url} 

\newif\ifarxiv
\arxivtrue

\ifarxiv
\else
\usepackage{lineno}
\fi

\begin{document}

\begin{frontmatter}



\title{VIFSS: View-Invariant and Figure Skating-Specific Pose Representation Learning for Temporal Action Segmentation}

\ifarxiv
\author[1]{Ryota Tanaka}
\ead{tanaka.ryota@g.sp.m.is.nagoya-u.ac.jp}

\author[1]{Tomohiro Suzuki}
\ead{suzuki.tomohiro@g.sp.m.is.nagoya-u.ac.jp}

\author[1,2]{Keisuke Fujii\corref{cor1}}
\ead{fujii@i.nagoya-u.ac.jp}
\cortext[cor1]{Corresponding author}

\affiliation[1]{
            organization={Graduate School of Informatics, Nagoya University},
            addressline={Chikusa-ku}, 
            city={Nagoya},
            postcode={464-8601}, 
            state={Aichi},
            country={Japan}}

\affiliation[2]{
            organization={RIKEN Center for Advanced Intelligence Project},
            addressline={1-5 Yamadaoka}, 
            city={Suita},
            postcode={565-0871}, 
            state={Osaka},
            country={Japan}}
\else
\author[1]{Anonymous}
\fi
\begin{abstract}
Understanding human actions from videos plays a critical role across various domains, including sports analytics. In figure skating, accurately recognizing the type and timing of jumps a skater performs is essential for objective performance evaluation. However, this task typically requires expert-level knowledge due to the fine-grained and complex nature of jump procedures. While recent approaches have attempted to automate this task using Temporal Action Segmentation (TAS), there are two major limitations to TAS for figure skating: the annotated data is insufficient, and existing methods do not account for the inherent three-dimensional aspects and procedural structure of jump actions.
In this work, we propose a new TAS framework for figure skating jumps that explicitly incorporates both the three-dimensional nature and the semantic procedure of jump movements. First, we propose a novel View-Invariant, Figure Skating-Specific pose representation learning approach (VIFSS) that combines contrastive learning as pre-training and action classification as fine-tuning. For view-invariant contrastive pre-training, we construct FS-Jump3D, the first publicly available 3D pose dataset specialized for figure skating jumps. Second, we introduce a fine-grained annotation scheme that marks the ``entry (preparation)'' and ``landing'' phases, enabling TAS models to learn the procedural structure of jumps.
Extensive experiments demonstrate the effectiveness of our framework. Our method achieves over 92\% F1@50 on \textit{element-level} TAS, which requires recognizing both jump types and rotation levels. Furthermore, we show that view-invariant contrastive pre-training is particularly effective when fine-tuning data is limited, highlighting the practicality of our approach in real-world scenarios. 
\end{abstract}

\begin{graphicalabstract}
\includegraphics[width=\textwidth]{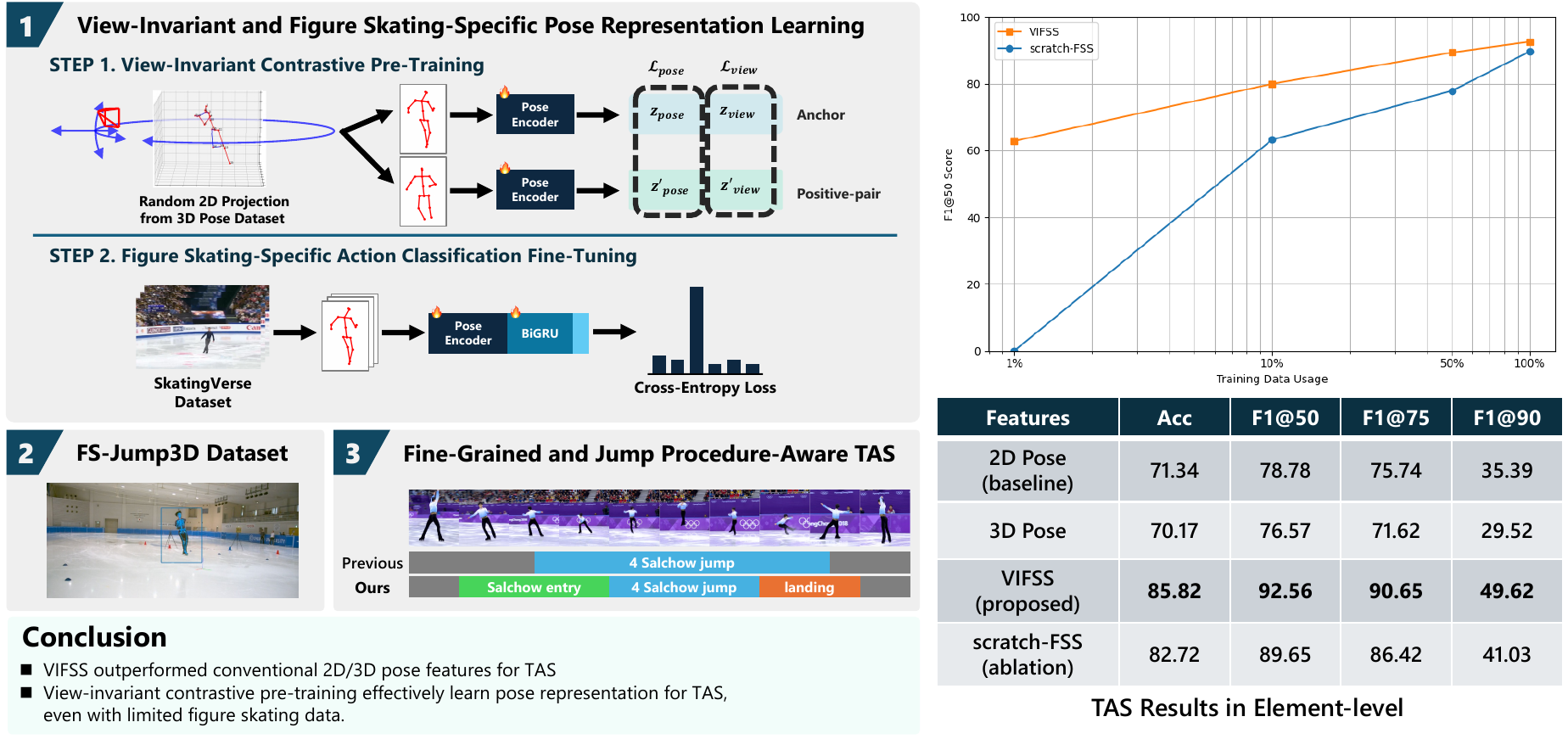}
\end{graphicalabstract}

\begin{highlights}
\item We introduce a new framework for temporal action segmentation using VIFSS (View-Invariant and Figure Skating-Specific) pose representations.
\item We construct FS-Jump3D, the first publicly available 3D pose dataset for figure skating jumps.
\item We propose a fine-grained annotation strategy to capture precise jump procedures.
\item Our VIFSS approach achieves superior performance on figure skating TAS tasks over 2D pose and 3D pose features.
\end{highlights}


\end{frontmatter}

\ifarxiv
\else
\linenumbers
\fi

\section{Introduction}\label{Introduction}
Understanding human actions from videos is a fundamental task with wide-ranging applications, from sports~\citep{suzuki2024automatic, tanaka2023automatic_mmsports} to autonomous driving~\citep{action_recog_pedestrian}, surveillance~\citep{action_recog_secure}, and elderly care~\citep{action_recog_nursing}. In the sports domain, video data plays a critical role in applications such as replay judgment~\citep{var}, automatic judgment~\citep{suzuki2024automatic, tanaka2023automatic_mmsports}, and providing feedback during training~\citep{sports_with_ai}. In particular, figure skating presents a unique challenge due to the increasing technical complexity of jump elements, making video utilization indispensable.

Currently, figure skating judging relies heavily on manual annotation by technical specialists and replay operators, who identify and record jump types and timings during performances, respectively. This process requires expert knowledge and a non-negligible amount of time and effort. To address these challenges, this study proposes an automated pipeline system that recognizes the type and timing of figure skating jumps from broadcast videos by Temporal Action Segmentation (TAS).

TAS~\citep{tas_analysis} aims to segment untrimmed video into frame-wise action labels. Existing benchmark datasets for TAS mainly consist of video recordings of procedural activities, such as cooking~\citep{breakfast, 50salads, epic-kitchen, gtea}, or assembling furniture or toys~\citep{ego-procel, assembly101, ikea_asm, meccano}. These datasets annotate fine-grained actions at the frame level, e.g., ``take cup'' and ``open fridge'', and TAS models are trained to understand these procedural transitions. A widely used and successful approach divides the TAS task into two stages: extracting temporal features and segmenting the actions based on those features~\citep{fact, mstcn, mstcn++}. The quality of the extracted features greatly influences segmentation performance, making feature selection a key consideration.

In standard TAS tasks, visual cues from objects or tools often help recognize actions. Consequently, image-based features such as I3D~\citep{i3d}, extracted using 3D-CNNs, are frequently adopted. However, previous studies on understanding figure skating actions~\citep{mcfs, vpd} have suggested that 2D poses or embedded pose representations may be more informative than image-based features. This reflects the nature of figure skating as a motion-centric task, where recognizing human movement is more important than the appearance of objects or backgrounds.

\begin{figure*}[ht]
    \centering
    \includegraphics[width=0.95\linewidth]{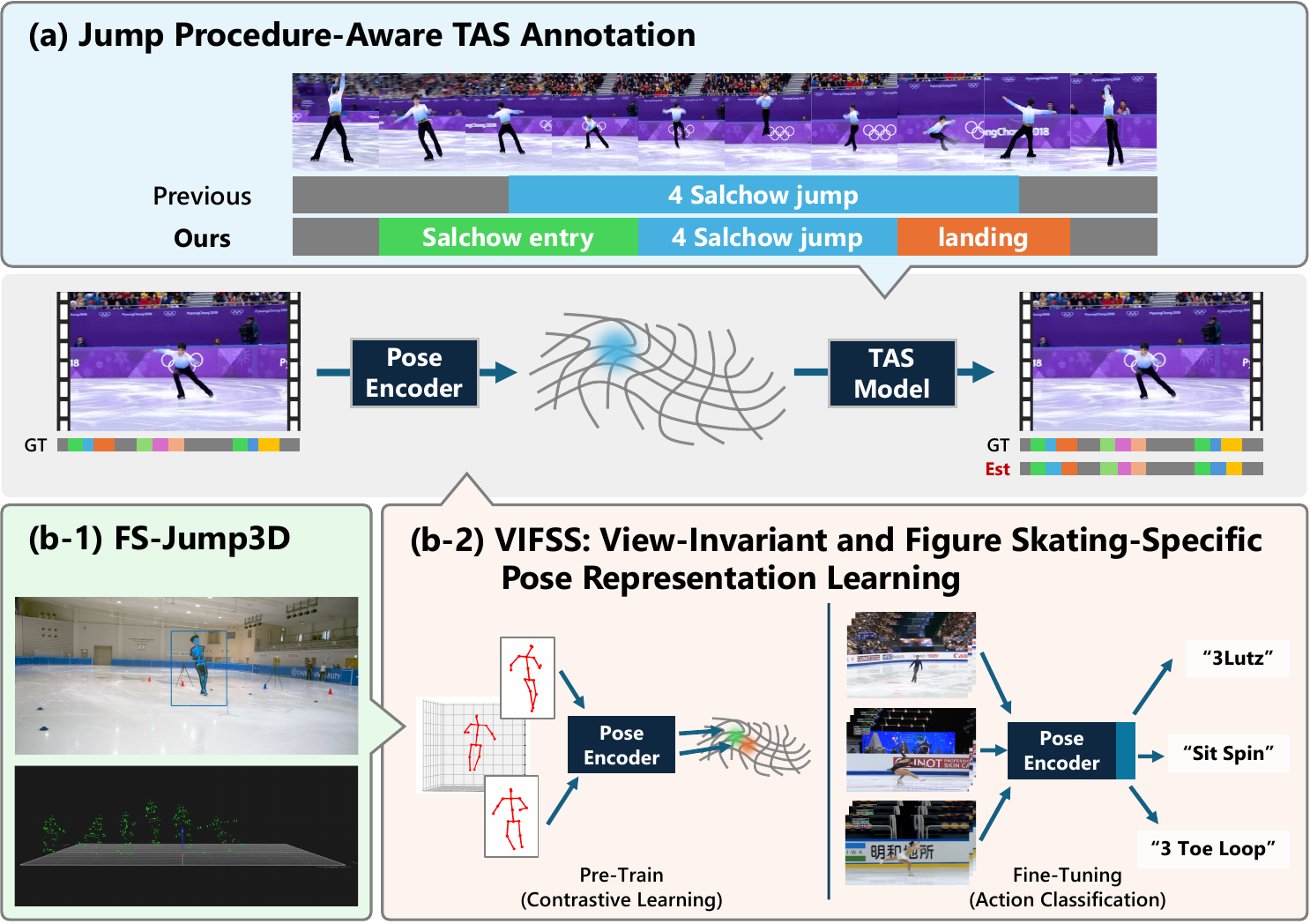}
    \caption{Overview of the proposed TAS framework.
    (a) We propose a novel annotation strategy that explicitly models the sequential structure of jump procedures.
    (b-1) To capture the three-dimensional and domain-specific characteristics of figure skating jumps, we construct FS-Jump3D, the first open 3D pose dataset dedicated to figure skating.
    (b-2) We introduce a two-stage pose representation learning framework, consisting of view-invariant contrastive pre-training and domain-specific fine-tuning via jump action classification.
    Finally, we evaluate the effectiveness of the learned pose embeddings as input to TAS models, combined with the proposed annotation scheme.}
    \label{fig:overview}
\end{figure*}

Nevertheless, figure skating actions are inherently three-dimensional and captured from a variety of viewpoints in broadcast footage. Leveraging 3D pose information to construct view-invariant pose representations offers promising potential for this task. However, no prior work has examined the effectiveness of 3D pose-based representations for TAS. Furthermore, existing TAS studies on figure skating~\citep{mcfs, skatingverse}, as illustrated in Figure~\ref{fig:overview} (a), do not account for entry (preparation) or landing phases in the annotation of jumps. Since TAS models are designed to learn sequential action procedures, explicitly labeling the full jump procedure, including the transition from entry to landing, may contribute to effective learning.

In this paper, we propose a new TAS framework that utilizes a View-Invariant and Figure Skating-Specific (VIFSS) pose representation learning method, which captures the three-dimensional nature of figure skating movements, as illustrated in Figure~\ref{fig:overview}.
First, this study addresses the above issues by proposing a new annotation strategy for figure skating TAS tasks that incorporates detailed jump procedures. We also investigate the utility of view-invariant and figure skating-specific 3D pose representations as input for the TAS model.

To address the challenges of incorporating the three-dimensional nature of figure skating motions into TAS, we introduce VIFSS that utilizes pose embedding features instead of directly using estimated 2D or 3D joint coordinates. While raw joint coordinates can represent motion geometrically, they often lack robustness to viewpoint variations and may hinder generalization across diverse domains. To overcome these limitations, we propose a two-stage learning approach. First, we pre-train a pose encoder using contrastive learning on 3D pose datasets to acquire view-invariant representations. Then, we fine-tune the encoder via action classification tailored to the figure skating domain. By combining view-invariant pre-training with domain-specific fine-tuning, our method enables more effective and flexible representation learning for TAS in figure skating, particularly under conditions of limited annotation and diverse viewpoints.

Although contrastive learning for pose representation has the advantage of not requiring explicit 3D pose datasets, relying only on synchronized multi-view videos, existing 3D pose datasets can further accelerate and enhance the learning process. However, most 3D pose datasets mainly focus on everyday activities such as walking or talking~\citep{h36m, panoptic, 3dpw, humaneva}, while sports-specific 3D pose datasets~\citep{sportspose, aspset, suzuki2025athleticspose} remain limited. Moreover, to the best of our knowledge, there is no publicly available 3D pose dataset capturing figure skating movements. To address this gap, we construct FS-Jump3D, a first open-source 3D pose dataset for figure skating jumps, using markerless motion capture. By integrating FS-Jump3D with other 3D pose datasets~\citep{h36m, 3dhp, aist++} within a unified training framework, we enable efficient learning of pose representations for the complex and dynamic motions of figure skating.

The aim of this study is to enable TAS for figure skating jumps with fine-grained detection of take-off and landing timings.

The key contributions of this work are as follows:
\begin{enumerate}
    \item We propose a two-stage learning framework called VIFSS for TAS, which combines contrastive pre-training with action classification fine-tuning to achieve view-invariant and domain-specialized embeddings for figure skating.
    \item We construct two datasets essential for learning the characteristics of figure skating: (i) FS-Jump3D, a 3D pose dataset of figure skating jumps, and (ii) a fine-grained TAS annotation set that captures jump procedures and the precise timings of take-off and landing.
    \item We conduct comprehensive experiments to validate the effectiveness of the proposed approach on figure skating TAS tasks.
\end{enumerate}

The FS-Jump3D dataset is publicly available to support research in figure skating and the analysis of complex sports motions. We also release the jump procedure-aware TAS annotations and the full project code for the proposed two-stage pose representation learning framework. We extend our previous conference workshop paper
\ifarxiv
~\citep{tanaka2024_mmsports} 
\else
(anonymized)
\fi 
by enhancing TAS performance through a novel VIFSS approach and conducting comprehensive experiments with detailed analysis.



\section{Related work}\label{Related work}
\subsection{Video Understanding in Sports}
In all types of sports, it is essential to quantitatively evaluate athletes' performance. Accordingly, various research tasks have been proposed to understand human motion in sports videos, including action prediction~\citep{soccer_prediction_2022, soccer_prediction_2024, badmiinton_prediction}, action recognition (classification)~\citep{soccer_var, fencenet, fitness_recognition, basketball_recognition, hockey_recognition_2023, hockey_recognition_2024, badminton_recognition, rugby_recognition}, temporal action detection~\citep{rugby_detection, e2e-spot, soccer_spotting_2023, soccer_spotting_2024, t-deed}, action quality assessment~\citep{aqa, aqa7, mtl-aqa, aqa_2024}, and TAS~\citep{mcfs, skatingverse}. 

Traditional approaches to these tasks have leveraged image features extracted through convolutional neural networks or optical flow~\citep{i3d, soccernet, soccer_i3d, baseball_i3d_1, baseball_i3d_2}. In recent years, however, methods based on object detection~\citep{yolo, yolov8}, object tracking~\citep{botsort, bytetrack}, and 2D/3D pose estimation~\citep{autosoccerpose, fitness_recognition, badminton_recognition, badmiinton_prediction} have become mainstream for extracting spatiotemporal information from video. Unlike image-based features, these approaches are more robust to variations in filming conditions and camera angles, making them particularly useful in the sports domain, where collecting large-scale datasets is often challenging. 

Understanding human actions in sports involves recognizing various elements, including group-level behaviors~\citep{basketball_recognition, hockey_recognition_2023, hockey_recognition_2024}, individual-level movements with joint dynamics~\citep{suzuki2024automatic, tanaka2023automatic_mmsports}, and object trajectories such as balls or other equipment~\citep{bascketball_3d_ball_localization, basketball_spin, tabletennis_spin, soccer_2d_ball_localization, badminton_shuttle}. Consequently, advancements in technologies such as object detection, tracking, and pose estimation are expected to significantly improve the accuracy of element recognition and, in turn, enhance the overall performance of downstream video understanding tasks.

\subsection{Video Understanding in Figure Skating}
In the context of figure skating, a range of tasks has been studied, including action recognition~\citep{vpd, fsd-10, skatingverse}, temporal action detection~\citep{e2e-spot}, action quality assessment~\citep{aqa, fs_score, audio_visual_mlp}, and TAS~\citep{mcfs, skatingverse}. Since figure skating is a judged sport that evaluates both technical skills and artistic expression, many studies aim to automate the scoring process. A common approach is to train deep learning models to regress overall scores from performance videos and corresponding judge scores from past competitions~\citep{aqa, fs_score, audio_visual_mlp}. However, these end-to-end methods often lack transparency and interpretability, making it difficult to understand how the models assess technical proficiency or artistic quality.

Recent studies in automated figure skating scoring have therefore shifted their focus to modeling individual judging components. This includes jump type classification~\citep{fsd-10, vpd}, edge error detection\footnote{Edge error detection evaluates whether the skater uses the correct edge (i.e., blade tilt) during take-off in accordance with the rules. It mainly applies to flip and lutz jumps and affects the base value and execution score of the jump.}~\citep{tanaka2023automatic_gcce, tanaka2023automatic_mmsports}, and under-rotation detection~\citep{aqa_fs}. In addition, TAS, which assigns frame-wise action labels to untrimmed performance videos, has also been studied for recognizing jumps and spins in figure skating.

Unlike many other sports that emphasize group dynamics or object trajectories, figure skating relies heavily on analyzing the skater’s fine-grained, sequential movements. FSD-10~\citep{fsd-10} tackled a classification task of the 10 most frequent jump and spin types using optical flow and 2D pose estimation, showing that focusing on frames with large joint displacements improved model performance. MCFS~\citep{mcfs} addressed TAS for jumps and spins, using 2D pose sequences as input features to improve segmentation accuracy. These studies suggest that pose-based features, particularly those invariant to background or costume, are beneficial for figure skating video understanding.

While the utility of 2D pose features has been validated, the potential of view-invariant 3D pose-based embeddings remains underexplored. Moreover, jump procedure elements such as preparation and landing phases are crucial for recognizing jump type and timing. However, no existing dataset for figure skating TAS has explicitly annotated such procedures. In this work, we propose a novel annotation method that incorporates jump procedure elements and investigate the effectiveness of using view-invariant pose representations for TAS in figure skating.

\subsection{3D Pose Estimation and Pose Representation Learning}
View-invariant pose representation can be achieved through two major approaches: directly estimating 3D poses or learning latent embeddings of poses.

For 3D pose estimation, there are two common approaches: one that directly estimates 3D poses from images, and another that lifts estimated 2D poses to 3D, known as 2D-to-3D lifting. With recent advances in 2D pose estimation, the latter approach has received more attention, allowing researchers to focus solely on the 2D-to-3D extension. This approach has demonstrated strong performance in recent works. Methods such as SimpleBaseline~\citep{simplebaseline}, SemGCN~\citep{semgcn}, and JointFormer~\citep{jointformer} propose 3D pose estimation models based on 2D-to-3D lifting from a single frame. In contrast, MotionBERT~\citep{motionbert} and MotionAGFormer~\citep{motionagformer} incorporate temporal modeling from multiple consecutive frames for 2D-to-3D lifting. While these temporal models enable more accurate 3D pose estimation, they often suffer from slower processing speeds and are limited to fixed-length frame sequences as input.

In pose representation learning, which aims to embed poses into a latent space, most methods similarly take 2D poses as input. Pr-VIPE~\citep{pr-vipe2020, pr-vipe2022} introduces a triplet loss-based framework to learn a latent space that captures pose similarity in a view-invariant manner. The resulting pose embeddings, produced by the encoder from 2D pose inputs, are effective for downstream tasks such as action recognition and video alignment. CV-MIM~\citep{cvmim} proposes a contrastive learning approach based on mutual information maximization to disentangle pose-dependent and view-dependent representations from 2D poses. While this method excels at capturing both pose-specific and viewpoint-specific features, a limitation is that it does not explicitly use viewpoint labels during training. As a result, it remains unclear whether the learned view-dependent features accurately reflect viewpoint similarity across different views.

\subsection{3D Pose Datasets}
Most existing 3D human pose datasets, such as Human3.6M~\citep{h36m}, 3DPW~\citep{3dpw}, MPI-INF-3DHP~\citep{3dhp}, and HumanEva-I~\citep{humaneva}, rely on optical marker-based motion capture systems and primarily contain daily activities such as walking and waving. While these systems offer high-precision tracking, they can constrain natural body movements, making them less suitable for capturing the complex dynamics of sports.

To address this limitation, several recent datasets have adopted markerless motion capture systems to record more dynamic activities, including soccer kicks and baseball pitches~\citep{sportspose}, jumping and catching~\citep{aspset}, and dance performances~\citep{aist++}. These datasets have broadened the applicability of 3D pose estimation models to high-motion scenarios that were not represented in conventional datasets.

Despite these advances, no public dataset currently exists for figure skating—a sport performed exclusively in the unique environment of an ice rink. On ice, skaters harness inertial and centrifugal forces to execute highly complex and dynamic movements that differ significantly from those in other sports. Existing datasets are therefore insufficient for modeling or estimating figure skating-specific poses.

To fill this gap, we introduce FS-Jump3D, the first 3D pose dataset dedicated to figure skating jumps. Captured on ice using a hardware-synchronized multi-camera setup and a markerless motion capture system, FS-Jump3D provides high-fidelity 3D jump motion data from expert skaters. For a detailed comparison of dataset characteristics, please refer to Table 1 in the previous conference paper~
\ifarxiv
\citep{tanaka2024_mmsports}.
\else
(anonymized).
\fi



\section{Methods}\label{Methods}
This study aims to perform TAS of figure skating jumps using VIFSS (View-Invariant and Figure Skating-Specific) pose representations as input features. Prior research~\citep{mcfs} has demonstrated that pose-based methods using 2D keypoints outperform image-based approaches, as they are less sensitive to variations in appearance, such as costumes or background. However, figure skating motions are inherently three-dimensional, and 2D poses, being projections of 3D body configurations onto the image plane, can vary significantly depending on the camera viewpoint. The direct 3D pose-based approaches face two key challenges: (1) TAS performance is highly sensitive to the quality of 3D pose estimation, and (2) simple coordinate-based representations may lack the capacity to capture complex motion dynamics.

\begin{figure*}[t]
    \vspace{10pt}
    \centering
    \includegraphics[width=0.95\linewidth]{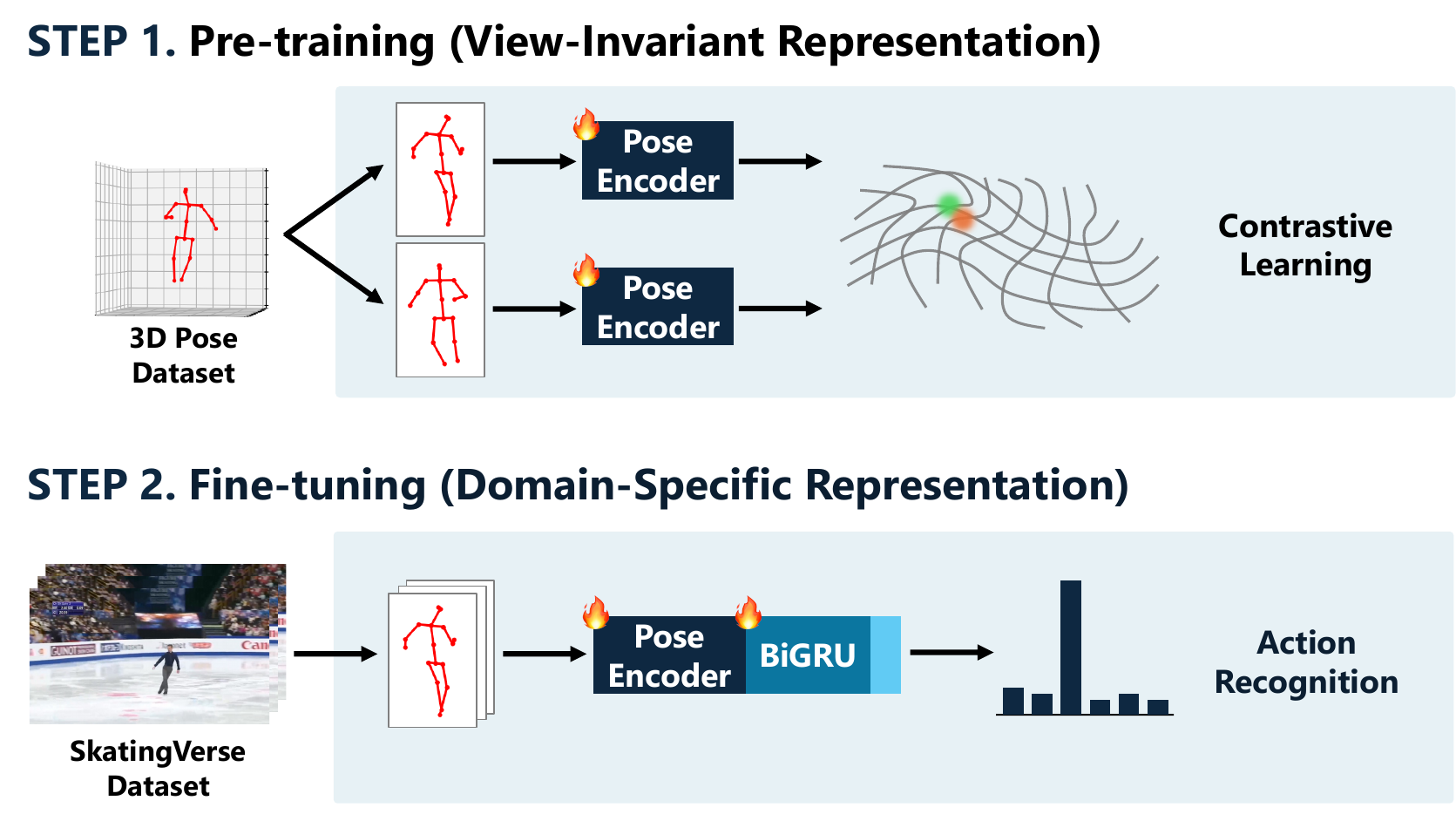}
    \caption{Overview of the proposed pose representation learning pipeline tailored for figure skating. In the first stage, contrastive learning is applied to a 3D pose dataset (FS-Jump3D and~\citep{h36m, 3dhp, aist++}) to pre-train a pose encoder that maps similar poses from different viewpoints close in the latent space. In the second stage, the pre-trained encoder is fine-tuned on the SkatingVerse~\cite {skatingverse} action classification dataset to specialize the pose representations for figure skating movements.}
    \label{fig:pose_embedding_learning}
    \vspace{30pt}
\end{figure*}

To address these challenges, we propose a method that employs contrastive learning with a 3D pose dataset to pre-train a pose encoder that extracts view-invariant pose embeddings. The pre-trained encoder is subsequently fine-tuned on a figure skating-specific action classification task, and the resulting pose embeddings serve as input features for the downstream TAS task. An overview of the proposed pose representation learning pipeline is illustrated in Figure~\ref{fig:pose_embedding_learning}. 

View-invariant pose representation learning via contrastive learning requires a suitable 3D pose dataset. However, most existing datasets~\citep{h36m,3dhp,3dpw,humaneva} focus on everyday activities and lack the domain-specific complex and dynamic motions of figure skating, which are driven by inertial and centrifugal forces. To the best of our knowledge, there are no publicly available 3D pose datasets tailored for figure skating. To address this limitation, we introduce FS-Jump3D, a 3D pose dataset capturing elite-level figure skating jumps, including triple rotations, recorded on ice using markerless motion capture.

We further propose a novel annotation strategy for TAS that considers the procedural structure of figure skating jumps. Since TAS models are typically designed to learn the procedures of human actions, annotating the preparation and landing phases of figure skating jumps can guide the model to better capture motion sequences of jumps.

The proposed TAS framework using VIFSS pose features is built upon the newly constructed FS-Jump3D dataset and a novel annotation scheme that accounts for the procedural structure of figure skating jumps. Accordingly, Section~\ref{sec:fs-jump3d} introduces the FS-Jump3D dataset. Section~\ref{sec:proposed_anntation} then presents our jump-specific TAS annotation method. Finally, Section~\ref{sec:VIFSS} describes the VIFSS pose features learned via a two-stage framework consisting of view-invariant contrastive pre-training and figure skating-specific fine-tuning.

\subsection{FS-Jump3D Dataset}\label{sec:fs-jump3d}
To capture the complex and dynamic motions of figure skating jumps, we construct a 3D pose dataset, FS-Jump3D, using a markerless motion capture system (Theia3D, Theia). Data collection is performed using 12 hardware-synchronized motion capture cameras (Miqus Video, Qualisys) installed around the ice skating rink. This configuration enables wide-area coverage, capturing not only the airborne phase of jumps but also the entry (preparation) and landing motions essential for TAS tasks. The markerless setup allows skaters to perform high-difficulty jumps, including triple jumps, without movement restrictions, while maintaining millimeter-level accuracy. We recorded jump data from four experienced figure skaters (referred to as Skaters A to D), each performing 10 trials for all six types of jumps. As a result, the FS-Jump3D dataset comprises 253 jump sequences, each containing synchronized footage from 12 camera views and the corresponding 3D pose data. Each pose consists of the 3D coordinates of 83 joints (head: 16, torso: 16, arms: 30, legs: 34), enabling detailed biomechanical analysis. For additional details on FS-Jump3D, we refer readers to the previous  paper
\ifarxiv
~\citep{tanaka2024_mmsports}.
\else
.
\fi

\subsection{Fine-Grained and Jump Procedure-Aware Annotation}\label{sec:proposed_anntation}
For the TAS dataset, we use broadcast footage of men's and women's short programs from the Olympic Games and World Championships, as adopted in prior work on jump classification in figure skating~\citep{vpd} and action spotting~\citep{e2e-spot}. This dataset comprises a total of 371 broadcast videos.

A limitation of previous studies on TAS in figure skating~\citep{mcfs,skatingverse} is that their annotation strategy does not consider the procedural phases of a jump (i.e., preparation and landing movements). Specifically, in these TAS datasets, only a single action label (e.g., ``2Axel'') is annotated for a few frames before take-off and after landing. Such annotations obscure the exact timing of take-off and landing, and lack the information to facilitate the TAS model's understanding of the sequential structure of jump motions. In figure skating, each type of jump has a distinctive preparatory movement to generate rotation and elevation. This makes the preparation phase essential for jump classification. Likewise, landing movements provide important cues for detecting the moment of landing.

To address this issue, we propose a novel annotation scheme that segments a figure skating jump into three phases:

\begin{enumerate}
    \item \textbf{entry:} The preparatory movements leading up to take-off.
    \item \textbf{jump:} The airborne phase from take-off to landing.
    \item \textbf{landing:} The movements immediately after landing.
\end{enumerate}

The entry phase includes frames starting from three steps before take-off, where a ``step'' is defined by changes such as turns or the skating leg switch. The landing phase begins at the frame where the skate blade first contacts the ice and continues while the skater maintains a sustained glide using the back outside edge\footnote{A proper landing posture in figure skating is characterized by a stable and flowing exit on the back outside edge.}. The entry labels are unique for each of the six jump types (e.g., ``Axel entry,'' ``Salchow entry''), whereas the landing label is common across all jump types.

Following prior work~\citep{mcfs,skatingverse}, we define a hierarchical label structure to control task difficulty in figure skating TAS, consisting of \textit{Set-level} and \textit{Element-level} annotations. At the Set-level, each jump is assigned one of six labels based on its type (e.g., ``Axel,'' ``Salchow''). The Element-level provides finer granularity with 23 labels that consider both jump type and the number of rotations (e.g., ``3 Axel,'' ``4 Salchow''). The Element-level task is more challenging, as it requires recognizing not only the jump type but also the number of rotations.

Combining the jump phase labels with the hierarchical label structure results in a total of 13 action labels at the Set-level and 30 labels at the Element-level. Frames not associated with any jumps (e.g., spins, step sequences) are annotated with the label ``NONE''.

\subsection{View-Invariant and Figure Skating-Specific Pose Representation Learning}\label{sec:VIFSS}
Prior work in figure skating action recognition~\citep{vpd} and TAS~\citep{mcfs} has demonstrated that using estimated 2D human poses as input leads to better performance than raw image features such as those extracted by I3D~\citep{i3d}. While 2D poses are robust to variations in background and costume, they are sensitive to changes in camera viewpoint and angle.
To mitigate this issue, the previous work
~\citep{tanaka2024_mmsports}
proposed a method that utilizes estimated 3D pose coordinates as input for TAS in figure skating. While this approach showed some performance gains (at Set-level), the accuracy of 3D pose estimation remains a bottleneck, implying that simple 3D coordinate-based representations are insufficient for fully capturing the complexity of figure skating motions.

To overcome these challenges, we propose a view-invariant and domain-specific pose encoder, using a 3D pose dataset for pre-training and an action classification dataset for fine-tuning, as illustrated in Figure~\ref{fig:pose_embedding_learning}. The training of this pose encoder consists of two main stages:

\begin{enumerate}
    \item \textbf{Pre-training with contrastive learning} to obtain view-invariant pose embeddings.
    \item \textbf{Fine-tuning on figure skating action classification} to adapt the pose encoder for figure skating-specific motions.
\end{enumerate}

In the pre-training stage, we adopt a contrastive learning framework that takes pairs of 2D poses captured of the same underlying 3D pose from different viewpoints. The goal is to learn pose embeddings that reflect multi-view similarity. In the fine-tuning stage, we optimize the pre-trained pose encoder for the figure skating action classification task, enhancing the encoder's ability to capture figure skating motions.

The following subsections provide a detailed description of the methodology for learning view-invariant and domain-specific pose embeddings proposed in this work.

\subsubsection{Pre-training a Pose Encoder via Contrastive Learning}
Inspired by prior work such as Pr-VIPE~\citep{pr-vipe2020,pr-vipe2022} and CV-MIM~\citep{cvmim}, we adopt a contrastive learning framework to learn view-invariant pose representations. In contrastive learning, as exemplified by SimCLR~\citep{simclr} and BYOL~\citep{byol}, different augmentations are applied to the same sample to create an “anchor” and a corresponding “positive” that share semantic similarity. The encoder is trained to map these pairs to nearby points in the embedding space. While image-based tasks typically use augmentations like masking, rotation, flipping, or jittering, we adapt this strategy for 3D human pose data. Specifically, we generate multiple 2D projections of 3D poses using virtual cameras at random viewpoints. These multi-view 2D poses are treated as anchor-positive pairs for training.

\paragraph{Pose Encoder and 3D Pose Datasets}
For the pose encoder, we adopt Jointformer~\citep{jointformer}, a transformer-based 3D human pose estimator designed for the per-frame 2D-to-3D lifting. In addition to our FS-Jump3D dataset, we incorporate several major 3D human pose datasets: Human3.6M~\citep{h36m}, MPI-INF-3DHP~\citep{3dhp}, and AIST++~\citep{aist++}. Human3.6M contains indoor recordings of everyday activities, while MPI-INF-3DHP includes both indoor daily motions and outdoor activities. AIST++ is a large-scale indoor dance motion dataset. Figure skating involves not only jumps but also a wide variety of motions. To improve the generalizability of the pose encoder and ensure it is adaptable to figure skating, we utilize a cross-dataset training strategy to expose the encoder to a diverse range of human poses.

\paragraph{Data Preprocessing}
As a preprocessing step for 3D pose datasets, we first standardized the keypoint definitions used to represent poses by aligning the number, spatial locations, and ordering of joint keypoints across datasets. Next, we apply alignment and normalization procedures to define consistent virtual camera directions for each 3D pose. Finally, we perform random augmentation during the loading of the training dataset to construct 2D pose pairs for contrastive learning.

\begin{figure*}[ht]
    \vspace{10pt}
    \centering
    \includegraphics[width=0.95\linewidth]{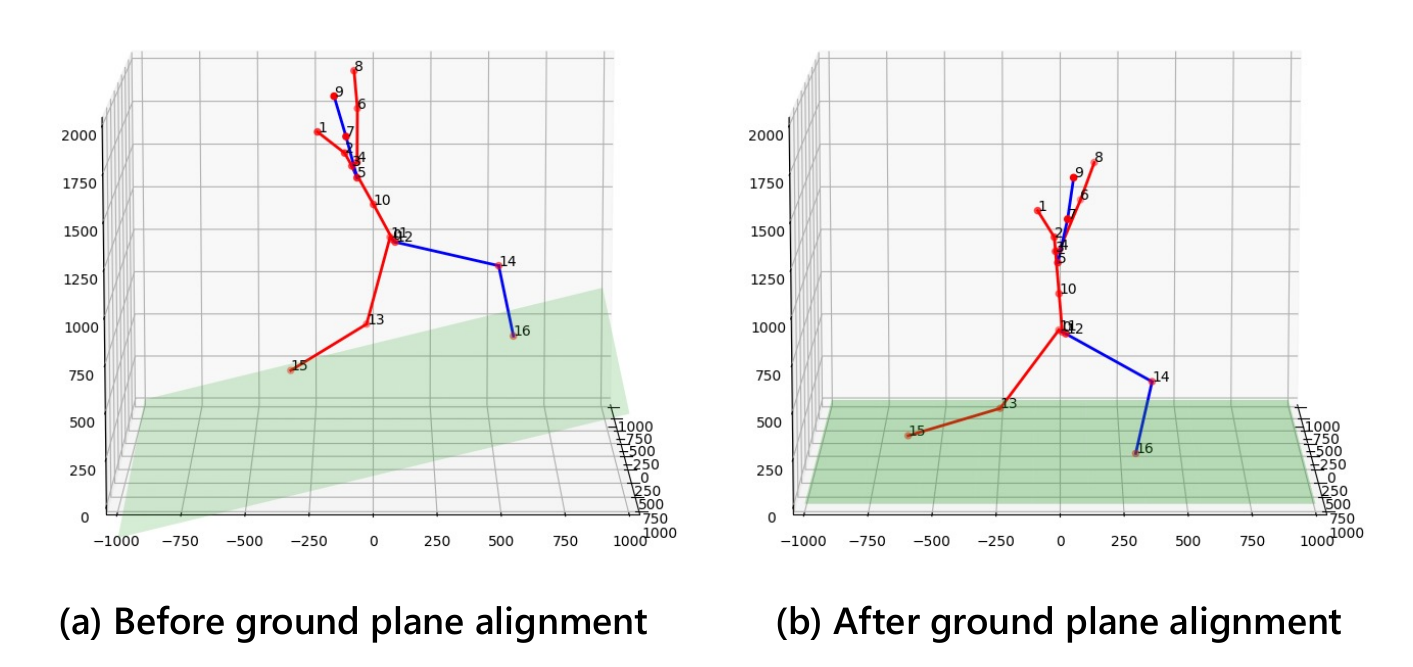}
    \caption{Ground plane alignment with RANSAC. We perform ground plane alignment across all 3D pose datasets such that the $z$-axis of the world coordinate system is aligned with the gravity direction, using plane detection based on RANSAC.}
    \label{fig:ground_plane_alignment}
    \vspace{20pt}
\end{figure*}

\begin{figure*}[ht]
    \vspace{10pt}
    \centering
    \includegraphics[width=0.95\linewidth]{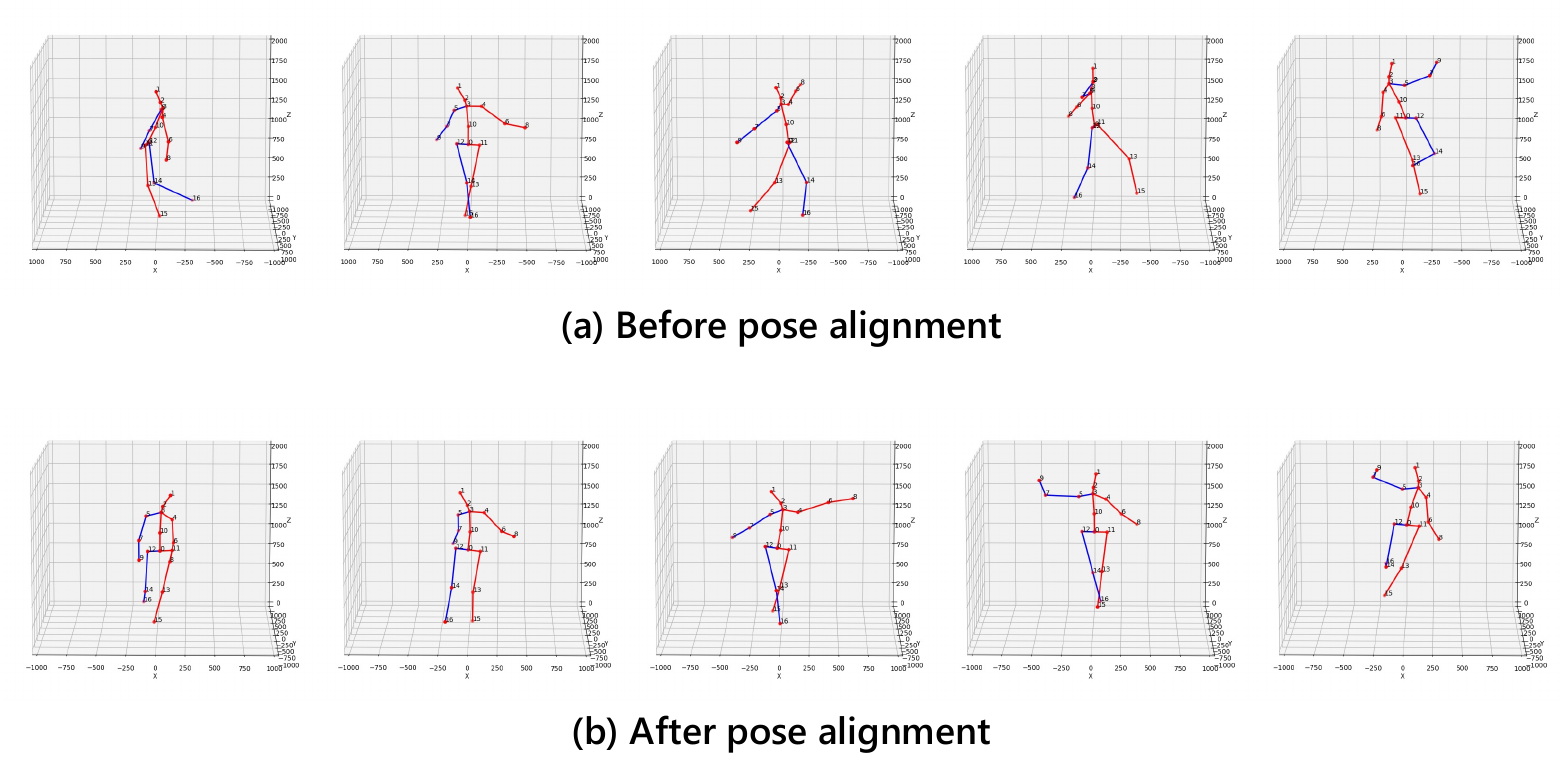}
    \caption{Pose alignment. For all 3D pose datasets, we align each pose in every frame so that it consistently faces the same direction throughout the sequence.}
    \label{fig:pose_alignment}
    \vspace{20pt}
\end{figure*}

For 3D pose alignment, we first align the gravity direction of the 3D pose to the global $z$-axis by estimating a ground plane via RANSAC~\citep{ransac} as illustrated in Figure~\ref{fig:ground_plane_alignment}. We detect the lowest $z$-coordinate in each pose and assume that 50\% of these lowest points correspond to actual contact points, while the remaining 50\% are treated as outliers. A ground plane is then estimated using RANSAC based on this assumption. The 3D pose is rotated so that its estimated ground plane aligns with the global $xy$-plane. Subsequently, we further rotate each pose around the $z$-axis so that all poses face the same direction, as shown in Figure~\ref{fig:pose_alignment}: the left hip joint is aligned to the positive $x$-axis and the right hip joint to the negative $x$-axis.

Normalization involves centering each pose at the mid-hip joint and rescaling the pose such that the sum of the distances from the mid-hip to the chest and from the chest to the neck equals 0.4. As a result, the centralized 3D joint coordinates are scaled to almost lie within an absolute value of 1.

Data augmentation includes random 2D projections using virtual cameras, horizontal flipping, jittering, and masking. The virtual camera is positioned relative to the origin-centered 3D pose, with its azimuth angle sampled uniformly from $\pm180^\circ$, elevation from $\pm30^\circ$, and distance from the range $[5, 10]$.

A 2D pose is generated by perspective projection from this virtual viewpoint. Horizontal flipping is performed by multiplying the $x$-coordinates of the 3D joints by $-1$ prior to projection. Jittering adds zero-mean Gaussian noise with a variance of $0.01$ to each 2D joint. Masking randomly sets 1\% of the 2D joint coordinates to $(0, 0)$ per frame. These augmentations enhance data diversity and promote learning of robust latent representations, improving resilience to noise in real-world 2D pose estimates.

\begin{figure*}[ht]
    \vspace{10pt}
    \centering
    \includegraphics[width=0.95\linewidth]{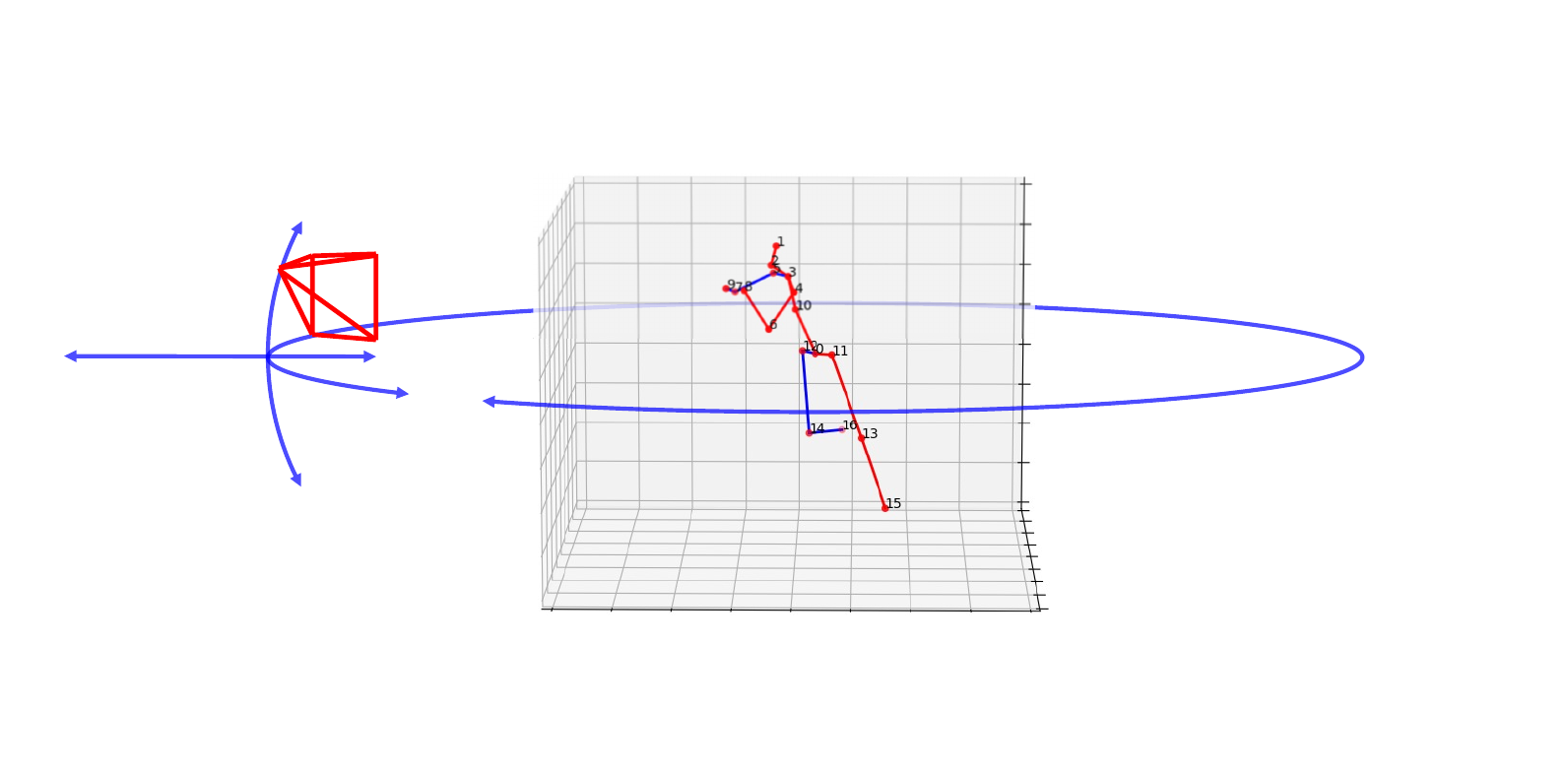}
    \caption{Random virtual camera arrangement. The virtual camera is positioned relative to the root joint (i.e., the origin) of the aligned and normalized 3D pose. The camera position is randomly sampled from a spherical coordinate range: the azimuth angle from $\pm180^\circ$, the elevation angle from $\pm30^\circ$, and the distance from the origin from the range $[5, 10]$. The virtual camera is always assumed to be oriented toward the root joint.}
    \label{fig:virtual_camera}
    \vspace{20pt}
\end{figure*}

\paragraph{Objective Function}
Inspired by CV-MIM~\citep{cvmim}, which learns view-invariant pose representations, we propose a simplified objective function that not only acquires such invariant features but also explicitly learns view-dependent features necessary for recognizing rotations in figure skating.

Let the 2D pose input be $x \in \mathbb{R}^{2 \times N}$ with $N$ joints, and the encoder output be an embedding $z \in \mathbb{R}^d$. The embedding is composed of a pose-invariant component $z_{\text{pose}} \in \mathbb{R}^{d_{\text{pose}}}$ and a view-dependent component $z_{\text{view}} \in \mathbb{R}^{d_{\text{view}}}$, such that $d = d_{\text{pose}} + d_{\text{view}}$.

Given an anchor pose and its positive pair, let $z$ and $z'$ be their respective embeddings. To enforce view-invariant representation learning, we define the pose loss $\mathcal{L}_{\text{pose}}$ using the Barlow Twins loss~\citep{barlowtwins}:
\[
\mathcal{L}_{\text{pose}} = \text{BarlowTwins}(z_{\text{pose}}, z'_{\text{pose}})
\]
The Barlow Twins loss reduces redundancy between anchor and positive embeddings, encouraging diverse and uncorrelated features without relying on negative samples, thus improving computational efficiency.

To learn view-dependent features, we define a view loss $\mathcal{L}_{\text{view}}$. Let $v_{\text{c}}$ and $v'_{\text{c}}$ denote unit vectors from the origin (mid-hip point) to the virtual cameras for the anchor and positive views, respectively. The loss is defined using cosine similarity and mean squared error (MSE) as:
\[
\mathcal{L}_{\text{view}} = \text{MSE}(\text{cossim}(z_{\text{view}}, z'_{\text{view}}), \text{cossim}(v_{\text{c}}, v'_{\text{c}}))
\]
This formulation incorporates virtual camera direction, consistently defined with respect to normalized and aligned 3D poses, into the learning objective. It encourages the model to capture features that are sensitive to viewpoint variations, which are typically neglected in conventional pose representation learning.

Finally, to prevent the learned embeddings from collapsing to trivial solutions and to encourage diversity in the embedding space, we introduce a regularization term $\mathcal{L}_{\text{R}}$. This term consists of two components: a variance loss that adjusts the output feature variance to a target value $\sigma_{\text{target}}^2$, and a KL divergence loss that regularizes the output distribution toward a uniform distribution:
\[
\mathcal{L}_{\text{R}} = \text{VarianceLoss}(z) + \text{VarianceLoss}(z') + \text{KLUniformLoss}(z) + \text{KLUniformLoss}(z')
\]
The two losses are defined as follows:
\[
\text{VarianceLoss}(z) = \frac{1}{d} \sum_{i=1}^{d} \left (\sigma_i^2(z) - \sigma_{\text{target}}^2 \right)^2
\]
\[
\text{KLUniformLoss}(z) = \frac{1}{d} \sum_{i=1}^{d} \left(z_i \log z_i + (1 - z_i) \log (1 - z_i) \right)
\]
In our experiments, we set $\sigma_{\text{target}}^2 = 1.0$ for the variance loss.

Based on the above, the overall loss function for contrastive learning of the pose encoder is defined as:
\[
\mathcal{L}_{\text{total}} = w_{\text{pose}} \cdot \mathcal{L}_{\text{pose}} + w_{\text{view}} \cdot \mathcal{L}_{\text{view}} + w_{\text{R}} \cdot \mathcal{L}_{\text{R}}
\]
where $w_{\text{pose}}$, $w_{\text{view}}$, and $w_{\text{R}}$ are the weights for each loss term. In our implementation, we set $w_{\text{pose}} = 1.0$, $w_{\text{view}} = 10.0$, and $w_{\text{R}} = 1.0$.

\subsubsection{Fine-Tuning the Pose Encoder via Action Classification}
After pre-training the pose encoder using contrastive learning to acquire view-invariant pose embeddings, we fine-tune the encoder to specialize in recognizing figure skating motions. This is achieved by training on a figure skating-specific action classification task.

\paragraph{Action Classification Dataset for Figure Skating}
For the action classification task, we use the SkatingVerse dataset~\citep{skatingverse}, which contains annotated video clips of figure skating jumps and spins categorized into 28 action classes. Of these, 23 classes correspond to combinations of six jump types (Axel, Salchow, Toe Loop, Loop, Flip, and Lutz) and four rotation levels (single to quadruple). The remaining five classes include four spin types (Camel Spin, Sit Spin, Upright Spin, and Other Spins), along with a ``NONE'' class for segments that do not contain jumps or spins.

The dataset provides annotations for 1,687 official figure skating videos, resulting in 19,993 training clips and 8,586 test clips, each trimmed and labeled according to its specific action.

\begin{figure*}[ht]
    \vspace{10pt}
    \centering
    \includegraphics[width=0.95\linewidth]{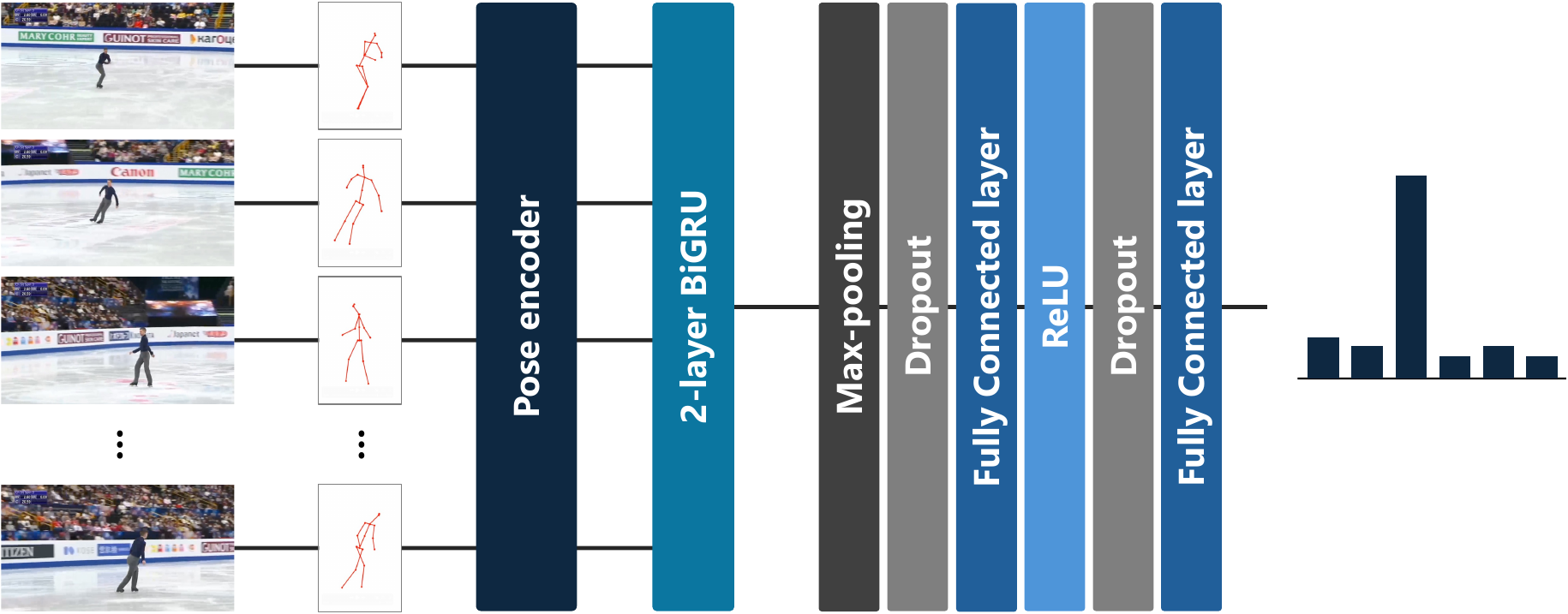}
    \caption{Overview of the action classification model using a two-layer BiGRU. The model fine-tunes the pose encoder through training on a figure skating action classification task.}
    \label{fig:action_recognition_model}
    \vspace{30pt}
\end{figure*}

\paragraph{Action Classification Model}
To fine-tune the pose encoder for figure skating action classification, we adopt a temporal modeling architecture inspired by a previous study~\citep{vpd}, which connects a BiGRU-based sequential model to the pre-trained pose encoder. The overall architecture is illustrated in Figure~\ref{fig:action_recognition_model}.

Given a sequence of 2D poses estimated from each video frame, the pose encoder outputs a sequence of pose embeddings. These embeddings are fed into a two-layer BiGRU, whose output is aggregated via temporal max pooling. The pooled feature is then passed through a fully connected layer with dropout, followed by a ReLU activation, another dropout layer, and a final fully connected layer to predict probabilities over 28 action classes.

The pose encoder is initialized with weights pre-trained using view-invariant contrastive learning. The entire model is fine-tuned using cross-entropy loss on the action classification task. This pre-training enables the model to efficiently learn representations tailored for figure skating motion understanding by fine-tuning.

\section{Experiments}\label{Experiments}
\subsection{Datasets}\label{datasets}
\subsubsection{FS-Jump3D Dataset}
The most distinctive feature of FS-Jump3D is its recording environment. While existing datasets are primarily collected under controlled indoor conditions, such as laboratories, and only a few are captured outdoors, FS-Jump3D was recorded in an ice skating rink, a highly specialized environment. Unlike standard ground conditions, the ice rink enables skaters to execute dynamic and complex actions that utilize inertia and centrifugal force. 

Another key feature is the use of markerless motion capture to record high-performance movements, including triple jumps. Figure skating jumps are extremely sensitive to environmental conditions and carry a risk of falls; thus, to avoid restricting skaters' natural movement, we adopted a markerless system. This approach allowed us to capture authentic jump data, including mistakes and falls, without compromising the difficulty level of the jumps.

Furthermore, FS-Jump3D was recorded using twelve dedicated motion capture cameras capable of hardware synchronization. Given the extremely rapid nature of jump movements in figure skating, hardware synchronization was essential to minimize inter-camera temporal misalignment and enable precise spatial reconstruction. The use of twelve cameras also provides sufficient multi-view coverage, which has the potential to mitigate severe occlusions commonly encountered in 2D pose estimation for figure skating.

\subsubsection{Fine-Grained and Jump Procedure-Aware TAS Annotation Statistics}
We constructed a TAS dataset of figure skating jumps, annotated using our proposed procedure-aware method. The dataset comprises 371 broadcast videos of short programs from international competitions, including the Winter Olympics (2010, 2014, 2018) and the World Figure Skating Championships (2017–2019), which cover both men's and women's events.

On average, each video contains 4,265 frames, of which approximately 382 frames (8.96\%) are annotated with action labels, corresponding to the interval from jump entry to landing. This sparsity highlights the inherent challenge of TAS in figure skating broadcasts, where annotated segments constitute only a small portion of the total video.

\begin{figure}[ht]
    \vspace{10pt}
    \centering
    \includegraphics[width=0.95\linewidth]{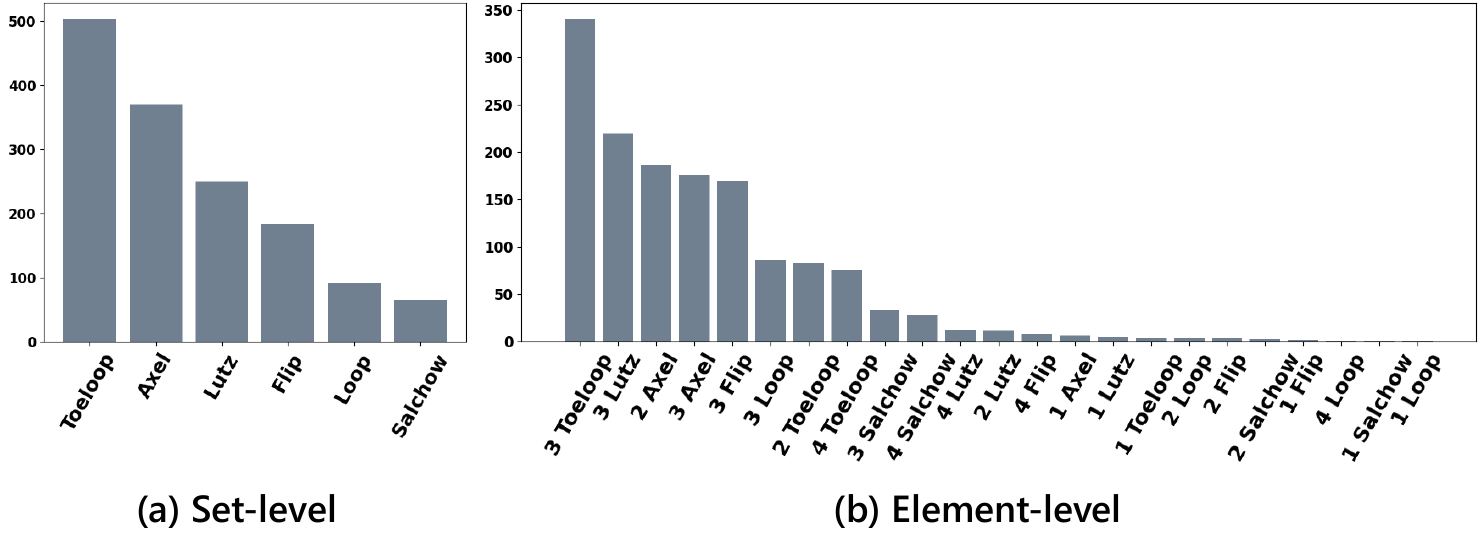}
    \caption{Annotation statistics of our TAS dataset for figure skating jumps. Both Set-level and Element-level labels exhibit significant class imbalance.}
    \label{fig:tas_anno_stats}
    \vspace{30pt}
\end{figure}

Figure~\ref{fig:tas_anno_stats} shows the distribution of jump types in the dataset. At the Set-level (a), toe loops are the most frequent, likely due to their occurrence as both standalone jumps and as the second jump in combinations. Axels are the second most frequent, reflecting their status as a required element in both men's and women's short programs.

At the element level (b), the most common jump is the triple toe loop, followed by the triple Lutz, double Axel, and triple Axel. The high frequency of Axels is consistent with their mandatory inclusion, with most women attempting a double Axel and most men attempting a triple Axel. The triple Lutz is also popular due to its high base value. In contrast, more difficult quadruple jumps and downgraded jumps (e.g., popped to singles or doubles) appear less frequently. These statistics reveal a significant class imbalance in element-level annotations and highlight the challenges of TAS in figure skating.

To evaluate the effectiveness of our procedure-aware annotation method, we define a baseline called \textit{Coarse Annotation} at the Set-level. In this baseline, all entry and landing labels are replaced with the ``None'' label, leaving only the jump action labels (e.g., ``Salchow'', ``Axel'') annotated frame by frame. This setup mimics the annotation style of MCFS~\citep{mcfs}, allowing for fair comparison while demonstrating the impact of incorporating detailed jump-procedural phases.

\subsection{Evaluation Protocol}
\subsubsection{Comparison of Pose Features for TAS Input}
To validate the effectiveness of the proposed approach, we conduct a series of evaluation experiments. First, we assess the impact of the proposed VIFSS pose features on TAS performance by comparing various types of input features. Additionally, we investigate the individual contributions of the FS-Jump3D dataset, the proposed annotation method, and our two-stage pose representation learning framework by evaluating TAS performance under different experimental settings.

To identify effective input representations for TAS in figure skating jumps, we compare the following four types of pose features, including our proposed method:

\begin{enumerate}
    \item 2D pose features (baseline)
    \item 3D pose features~\citep{tanaka2024_mmsports}
    \item VIFSS features (proposed)
    \item scratch-FSS features (ablation)
\end{enumerate}

As a baseline, we employ \textit{2D pose features}, which are commonly used in figure skating action recognition~\citep{mcfs}. These are 2D coordinates directly obtained via 2D pose estimation on each video frame. In our implementation, we utilize DWPose~\citep{dwpose}, pre-trained on the COCO-Wholebody dataset and publicly released by the authors, and normalize the obtained 2D joint coordinates.

The \textit{3D pose features} are generated by lifting the estimated 2D poses to 3D and applying view-invariant normalization, aligning all poses to face a consistent direction. We utilize MotionAGFormer~\citep{motionagformer} for 3D pose estimation, trained on the same dataset used to generate the proposed VIFSS pose features. Additionally, the 3D rotation angles used during pose alignment are normalized and concatenated to the aligned pose coordinates as auxiliary features, enabling the representation to retain global orientation cues that are beneficial for action recognition. For further details on the 3D pose feature processing pipeline, please refer to the preliminary conference paper~\citep{tanaka2024_mmsports}.

The proposed \textit{VIFSS features} are compared with the \textit{scratch-FSS features} used in our ablation study. Here, ``VIFSS'' indicates that the pose encoder is first pre-trained via contrastive learning to acquire view-invariant representations, followed by fine-tuning on a figure skating-specific action classification task. In contrast, the \textit{scratch-FSS features} are extracted from an encoder trained from scratch solely on the action classification task, without the view-invariant pre-training stage. This comparison allows us to quantify the contribution of the proposed pre-training strategy to the quality of the learned pose representations.

\subsubsection{Evaluation scheme of TAS}
We evaluate the effectiveness of various input feature representations for TAS, using our proposed fine-grained annotation that incorporates detailed jump procedures.

Following the experimental protocol of prior work~\citep{vpd}, all footage from the 2018 Winter Olympics and World Figure Skating Championships is designated as the test set. The remaining videos are used for training, with 20\% of the training data held out for validation. This split ensures that no footage from the same competition year appears in both the training and test sets, thereby enabling an evaluation of the model’s generalization capability to unseen video conditions, such as variations in camera placement, viewing angles, and video quality.

To assess the utility of the proposed pose-based features, we adopt FACT~\citep{fact} as the TAS model. FACT is a Transformer-based architecture that leverages cross-attention mechanisms to jointly learn frame-level and action-level representations. It has demonstrated state-of-the-art performance on several benchmark TAS datasets~\citep{breakfast,50salads,gtea,epic-kitchen}.

We follow standard evaluation protocols in the TAS literature~\citep{tas_analysis}, computing frame-wise accuracy and F1@\{10, 25, 50, 75, 90\} scores. These metrics are calculated over all action segments, excluding the ``entry'', ``landing'', and ``None'' labels. Frame-wise accuracy quantifies the proportion of correctly labeled frames. The F1@k metric considers a predicted segment correct if it overlaps with the corresponding ground-truth segment by at least \emph{k}\% of its length, and computes the harmonic mean of precision and recall at each overlap threshold.



\section{Results}\label{Results}
This section presents a comprehensive evaluation of our proposed method for TAS of figure skating jumps under various experimental settings.
In Section~\ref{sec:tas_set_vs_element}, we compare TAS performance between two annotation granularities: Set-level and Element-level. We then evaluate the effectiveness of our proposed VIFSS pose features against alternative feature types at both annotation levels. Detailed results for each level are discussed in Sections~\ref{sec:tas_set} and~\ref{sec:tas_element}.
Section~\ref{sec:tas_fsjump3d} investigates the impact of the FS-Jump3D dataset by comparing view-invariant pre-training with and without its inclusion.
In Section~\ref{sec:annotation_comparison}, we assess the benefit of our fine-grained, jump procedure-aware annotation scheme by contrasting it with coarse annotations, highlighting its role in improving the model’s understanding of jump dynamics.
Finally, Section~\ref{sec:ablation_for_pretraining} presents an ablation study quantifying the contribution of view-invariant pre-training under varying supervision levels, with particular emphasis on its advantages in low-data scenarios.

\subsection{Comparison Between Set-level and Element-level TAS Annotations}
\label{sec:tas_set_vs_element}
Tables~\ref{tab:tas_set_vs_element} report the TAS results for figure skating jumps using different input features, comparing the performance between Set-level and Element-level annotations. A comparison of Tables~\ref{tab:set_level} and~\ref{tab:element_level} reveals that segmentation at the Element-level is more challenging than at the Set-level. In both settings, the F1@90 scores are substantially lower than the scores for F1@10 to F1@75. In our broadcast video dataset, the average duration of a jump, from take-off to landing, was 16.25 frames. Under the F1@90 metric, a predicted segment is considered correct only if it overlaps with the ground-truth segment by approximately 15 frames, allowing for an error margin of merely 1--2 frames. In contrast, the F1@75 metric requires approximately 12 frames of overlap, allowing for an error margin of roughly four frames (i.e., two frames before and after the jump). Under this F1@75 metrics, our proposed VIFSS features employing two-stage pose representation learning achieve over 90\% in both Set-level and Element-level tasks. 

Our annotation scheme enables precise evaluation of take-off and landing timings. While F1@90 serves as an upper bound for model precision, the consistently high F1@75 scores demonstrate the practical effectiveness of the proposed representations.

\begin{table}[ht]
    \centering
    \begin{subtable}[t]{\textwidth}
        \centering
        \small
        \begin{tabular}{lcccccc}
        \hline
        Feature & Acc & F1@10 & F1@25 & F1@50 & F1@75 & F1@90 \\
        \hline
        2D pose (Baseline) & 78.55 & 85.12 & 84.93 & 84.17 & 81.52 & 35.83 \\
        3D pose~\citep{tanaka2024_mmsports} & 79.89 & 87.13 & 86.94 & 86.56 & 82.36 & 33.36 \\
        \textbf{VIFSS} & \textbf{89.91} & \textbf{95.44} & \textbf{95.44} & \textbf{94.68} & \textbf{93.16} & \textbf{51.71} \\
        scratch-FSS & 86.38 & 92.48 & 92.29 & 91.72 & 88.87 & 42.44 \\
        \hline
        \end{tabular}
        \caption{Set-level TAS results.}
        \label{tab:set_level}
    \end{subtable}

    \vspace{1em}

    \begin{subtable}[t]{\textwidth}
        \centering
        \small
        \begin{tabular}{lcccccc}
        \hline
        Feature & Acc & F1@10 & F1@25 & F1@50 & F1@75 & F1@90 \\
        \hline
        2D pose (Baseline) & 71.34 & 78.97 & 78.97 & 78.78 & 75.74 & 35.39 \\
        3D pose~\citep{tanaka2024_mmsports} & 70.17 & 77.71 & 77.33 & 76.57 & 71.62 & 29.52 \\
        \textbf{VIFSS} & \textbf{85.82} & \textbf{92.75} & \textbf{92.75} & \textbf{92.56} & \textbf{90.65} & \textbf{49.62} \\
        scratch-FSS & 82.72 & 89.65 & 89.65 & 89.65 & 86.42 & 41.03 \\
        \hline
        \end{tabular}
        \caption{Element-level TAS results.}
        \label{tab:element_level}
    \end{subtable}

    \caption{Comparison of TAS performance at different granularity levels of annotation.}
    \label{tab:tas_set_vs_element}
\end{table}

\subsection{Set-Level TAS Performance}
\label{sec:tas_set}
As shown in Table~\ref{tab:set_level}, both the VIFSS and scratch-FSS features outperform the baseline methods in Set-level TAS. Among these, the VIFSS features achieve the highest overall performance. These results indicate that the proposed view-invariant contrastive pre-training effectively mitigates the view-dependent nature of 2D pose features, thereby enhancing downstream action classification performance for figure skating-specific movements. Although the 3D pose features lag behind the proposed pose embeddings, they still surpass the baseline 2D pose features. Overall, these findings demonstrate that learning view-invariant pose representations, particularly the proposed VIFSS features, is effective for identifying both the type and timing of jumps at the Set level.

\subsection{Element-Level TAS Performance}
\label{sec:tas_element}
Table~\ref{tab:element_level} presents the TAS results at the Element-level for each type of input feature. Consistent with the Set-level findings, the pose embedding features, VIFSS and scratch-FSS features, achieve strong performance. Especially, the VIFSS features exceed 92\% at F1@50. In contrast, the 3D pose features perform worse than the baseline 2D pose features at the Element-level. This suggests that the proposed two-stage learning framework for pose embeddings is also effective for fine-grained Element-level segmentation. Meanwhile, the 3D pose features appear less suitable for tasks that require precise recognition of complex rotational motion.

The degradation in the performance of 3D pose features at the Element-level is likely due to limitations in the 3D pose estimation model itself. Specifically, MotionAGFormer~\citep{motionagformer}, the 3D pose estimator used in this study, is a temporal model that predicts 3D poses from sequences of 2D poses. However, differences in the temporal characteristics of the training and inference data likely affected its performance. The FS-Jump3D training dataset primarily consists of double jumps performed by subjects B and C, and triple jumps by subject A. In contrast, the broadcast videos used for inference mainly contain triple jumps and even quadruple jumps. Moreover, while the FS-Jump3D dataset was downsampled from 60fps to 30fps for training, the broadcast videos were recorded at 25fps. These discrepancies in the number of rotations and frame intervals between training and inference may have led to a decline in the model’s ability to recognize rotational motion.

\begin{figure}[ht]
    \centering
    \begin{subfigure}[b]{0.99\linewidth}
        \centering
        \includegraphics[width=\linewidth]{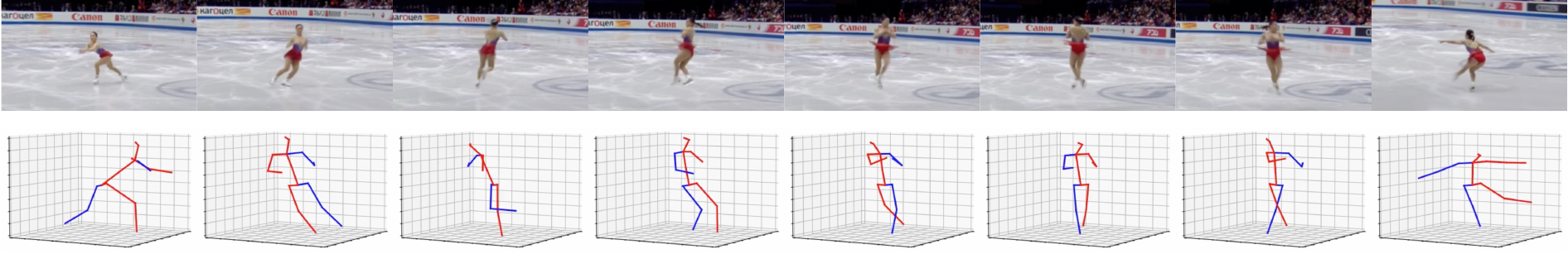}
        \caption{Successful 3D pose estimation of a double Axel in a broadcast video. The rotational motion is accurately reconstructed as a 3D pose sequence.}
        \label{fig:correct_pose3d_sequence}
    \end{subfigure}
    
    \vspace{10pt}
    
    \begin{subfigure}[b]{0.99\linewidth}
        \centering
        \includegraphics[width=\linewidth]{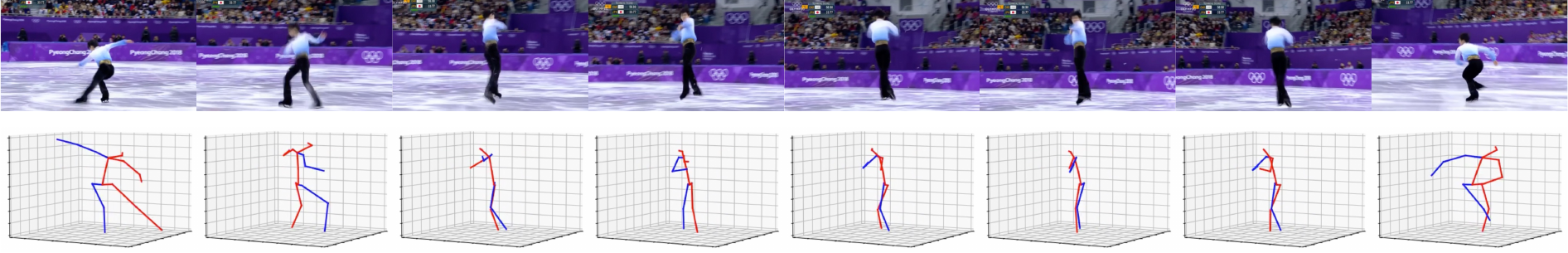}
        \caption{Failed 3D pose estimation of a quadruple toe loop in a broadcast video. The model fails to capture the rotational motion, likely due to the absence of similar high-speed rotations in the training data.}
        \label{fig:error_pose3d_sequence}
    \end{subfigure}
    
    \caption{Examples of 3D pose estimation for jump rotations in broadcast videos. (a) shows a successful case, while (b) illustrates a failure likely caused by limitations in training data diversity.}
    \label{fig:pose3d_sequence_comparison}
\end{figure}

Figures~\ref{fig:correct_pose3d_sequence} and~\ref{fig:error_pose3d_sequence} show examples of successful and failed 3D pose estimations of jump rotations in broadcast videos, respectively. Figure~\ref{fig:correct_pose3d_sequence} presents a double Axel, which was well-represented in the FS-Jump3D training set (performed by subjects A and C), and demonstrates accurate 3D pose estimation of the rotational motion. In contrast, Figure~\ref{fig:error_pose3d_sequence} shows a quadruple toe loop, which was not included in the training data, resulting in a failure to correctly capture the rotational motion. Similarly, failures were often observed for jumps with unusual arm positions, such as triple jumps performed with both arms raised above the head. These observations suggest that the performance of temporal 3D pose estimation models heavily depends on the diversity of the training 3D pose dataset and that the model's limitations become a bottleneck for subsequent TAS tasks.

In contrast, the pose embedding approach demonstrated robustness to such variations. Since the contrastive pre-training stage is designed to learn per-frame pose representations independent of temporal sequences, it is less affected by intra-class variability in jump rotations. Moreover, the fine-tuning stage does not rely on 3D pose annotations, which are often difficult to obtain, making it more practical for domain-specific applications. As a result, the embedding-based approach consistently achieved high performance even for the more demanding Element-level TAS task.

\subsection{Impact of FS-Jump3D on TAS Performance}
\label{sec:tas_fsjump3d}
Table~\ref{tab:set_level_wo_fs} compares the Set-level TAS performance when the FS-Jump3D dataset is included or excluded from the training process, using both 3D pose features and VIFSS features. The results show that excluding the FS-Jump3D dataset from training for both types of features leads to a degradation in segmentation performance. FS-Jump3D consists of figure skating jump actions captured in a unique environment, an ice rink, which differs significantly from conventional 3D pose datasets. These results indicate that FS-Jump3D plays a crucial role in accurately capturing jump motions in figure skating, underscoring its importance for the TAS task.

\begin{table*}[ht]
    \vspace{20pt}
    \centering
    \small
    \begin{tabular}{lcccccc}
    \hline
    Feature & Acc & F1@10 & F1@25 & F1@50 & F1@75 & F1@90 \\
    \hline
    3D pose~\citep{tanaka2024_mmsports} & 79.89 & 87.13 & 86.94 & 86.56 & 82.36 & 33.36 \\
    3D pose (w/o FS-Jump3D) & 75.76 & 83.08 & 82.89 & 82.51 & 77.57 & 31.18 \\
    \textbf{VIFSS} & \textbf{89.91} & \textbf{95.44} & \textbf{95.44} & \textbf{94.68} & \textbf{93.16} & \textbf{51.71} \\
    VIFSS (w/o FS-Jump3D) & 87.53 & 93.16 & 92.97 & 92.78 & 91.25 & 46.01 \\
    \hline
    \end{tabular}
    \caption{TAS results on the Set-level with and without the FS-Jump3D dataset.}
    \label{tab:set_level_wo_fs}
    \vspace{20pt}
\end{table*}

\subsection{Comparison Between Proposed and Coarse Annotations}
\label{sec:annotation_comparison}
The proposed annotation method in this study aims to incorporate knowledge about the procedural steps of jumps (entry and landing phases) into the TAS model, thereby helping the model to better recognize the type and timing of jumps. Table~\ref{tab:annotation_comparison} presents the results of validating the effectiveness of the proposed annotation on the Set-level TAS task. For comparison, a \textit{Coarse} annotation was created by replacing all labels related to entry and landing phases in the proposed annotations with a ``None'' label, leaving only the jump labels as valid action classes.  
The evaluation metric used was F1@50, which is the most commonly adopted metric for TAS.

The results suggest that introducing procedural steps through the proposed annotation leads to improved TAS performance. In addition to providing the model with a deeper understanding of jump procedures, the proposed annotation might also contribute to performance improvement by assigning meaningful labels to a larger number of frames. Specifically, in the coarse annotation, only 1.50\% of all frames are assigned action labels (jump only), whereas in the proposed annotation, 8.96\% of frames are labeled with action labels (entry, jump, or landing). This difference in the proportion of action labels suggests that providing meaningful frame-level annotations over a broader time span is an important factor in enhancing TAS performance for figure skating jumps.

\begin{table}[ht]
    \vspace{10pt}
    \centering
    \small
    \setlength{\tabcolsep}{10pt} 
    \renewcommand{\arraystretch}{1.3} 
    \begin{tabular*}{0.6\linewidth}{@{\extracolsep{\fill}}lcc}
    \hline
    Feature & Proposed & Coarse \\
    \hline
    2D pose (baseline) & \textbf{84.17} & 76.42 \\
    3D pose~\citep{tanaka2024_mmsports} & \textbf{86.56} & 72.78 \\
    VIFSS & \textbf{94.68} & 93.93 \\
    \hline
    \end{tabular*}
    \caption{Comparison between the proposed and coarse annotations based on F1@50 for the Set-level TAS.}
    \label{tab:annotation_comparison}
    \vspace{10pt}
\end{table}

\subsection{Effectiveness of View-Invariant Pre-training}
\label{sec:ablation_for_pretraining}
\begin{figure}[htbp]
    \centering
    \begin{subfigure}[b]{0.49\linewidth}
        \centering
        \includegraphics[width=\linewidth]{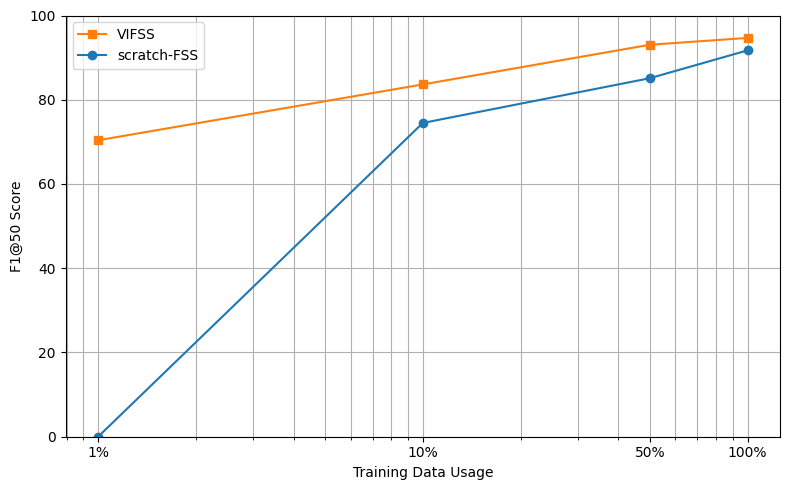}
        \caption{F1@50 at the Set-level with and without view-invariant pre-training.}
        \label{fig:pre-train_abration_set}
    \end{subfigure}
    \hfill
    \begin{subfigure}[b]{0.49\linewidth}
        \centering
        \includegraphics[width=\linewidth]{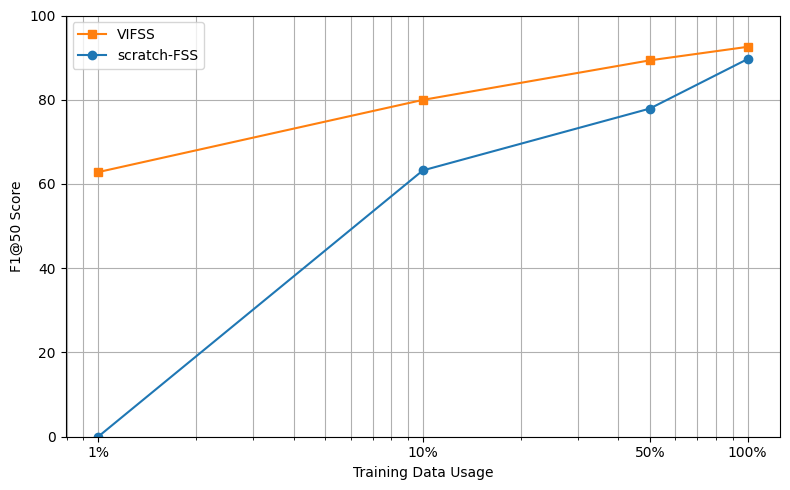}
        \caption{F1@50 at the Element-level with and without view-invariant pre-training.}
        \label{fig:pre-train_abration_element}
    \end{subfigure}
    
    \caption{Impact of view-invariant pose representation pre-training on TAS performance across different proportions of fine-tuning data. In both annotation levels, pre-training consistently improves TAS performance, with particularly significant gains observed under low-data conditions.}
    \label{fig:pre-train_abration}
\end{figure}

Figure~\ref{fig:pre-train_abration} shows the results of an ablation study evaluating the impact of pre-training in our two-stage learning framework for VIFSS features. We compared models with and without contrastive pre-training across both the Set-level and Element-level TAS tasks. For the second-stage fine-tuning, we varied the amount of action classification data to 100\%, 50\%, 10\%, and 1\%, and evaluated the performance using F1@50.

The results demonstrate that incorporating view-invariant contrastive pre-training consistently improves TAS performance at both levels. Notably, the benefit of pre-training becomes more pronounced as the amount of fine-tuning data decreases. In the extreme case of using only 1\% of the data, models without pre-training failed to learn meaningful pose embeddings, resulting in near-zero TAS performance. In contrast, models with pre-training achieved over 70\% F1@50 at the Set-level and over 60\% at the Element-level.

These findings highlight that view-invariant pre-training is particularly effective in low-data regimes. It enables efficient learning of domain-adaptive pose embeddings for figure skating, even when the annotated fine-tuning data is severely limited.

\section{Conclusion}
\label{Conclusion}
In this work, we proposed VIFSS pose features and demonstrated their effectiveness for TAS of figure skating jumps. We also introduced a novel annotation strategy that incorporates jump procedures and constructed FS-Jump3D, the first publicly available 3D pose dataset for figure skating, including triple jumps.

Extensive experiments showed that pose embeddings derived from a pose encoder pre-trained via contrastive learning and fine-tuned on a figure skating action classification task significantly improve TAS performance at both the Set-level and Element-level. These results suggest that learning latent pose representations with domain-specific supervision offers greater flexibility and accuracy than direct regression of 3D joint coordinates. Furthermore, comparisons with Coarse annotations confirmed the advantages of our procedure-aware annotations in providing richer and more informative supervision signals for TAS. Ablation studies also confirmed that our two-stage VIFSS learning approach powerfully supports domain adaptation, particularly in scenarios with limited annotation.

Building on the demonstrated efficacy of view-invariant contrastive pre-training and task-specific fine-tuning in figure skating, future work will extend our approach to TAS for other figure skating elements such as spins and step sequences. We also plan to explore its generalizability to TAS tasks in other sports domains, aiming to further validate the robustness and versatility of our proposed method.


\ifarxiv
\section*{Author contributions}
TS supported the creation and processing of the FS-Jump3D dataset and contributed to the manuscript revision. KF supervised this study and contributed to the manuscript revision. All authors read and approved the final manuscript.

\section*{Acknowledgments}
This work was funded by JSPS Grant Number 21H05300 and 23H03282, and JST PRESTO Grant Number JPMJPR20CA. The funder played no role in study design, data collection, analysis, interpretation of data, or the writing of this manuscript.

\section*{Data and Code Availability}
The TAS dataset annotation and source code for our VIFSS pose representation learning method will be made publicly available at \url{https://github.com/ryota-skating/VIFSS} upon publication.

\fi

\section*{Competing interests}
All authors declare no competing interests. 



\bibliographystyle{elsarticle-harv} 
\bibliography{ref}

\end{document}